\documentclass{article}
\usepackage{float}

\usepackage{PRIMEarxiv}
\usepackage{caption}

\usepackage[utf8]{inputenc} 
\usepackage[T1]{fontenc}    
\usepackage{hyperref}       
\usepackage{url}            
\usepackage{booktabs}       
\usepackage{amsfonts}       
\usepackage{nicefrac}       
\usepackage{microtype}      
\usepackage{lipsum}
\usepackage{fancyhdr}       
\usepackage{graphicx}       
\usepackage{amsmath}
\usepackage{comment}
\usepackage{graphicx}
\usepackage{subcaption}
\graphicspath{{media/}}     
\usepackage{authblk}
\title{Predicting Region of Interest in Human Visual Search Based on Statistical Texture and Gabor Features
}

\author[1]{Hongwei Lin}
\author[1]{Diego Andrade}
\author[1,2,3,**]{Mini Das}
\author[1,*]{Howard C. Gifford}

\affil[1]{Department of Biomedical Engineering, University of Houston, 3517 Cullen Blvd, Houston, TX 77204, USA}
\affil[2]{Department of Physics, University of Houston, 3507 Cullen Blvd, Houston, TX 77204, USA}
\affil[3]{Department of Electrical and Computer Engineering, University of Houston, 3517 Cullen Blvd, Houston, TX 77204, USA}
\affil[*]{\texttt{hgifford@central.uh.edu}}
\affil[**]{\texttt{mdas@central.uh.edu}}

\begin{document}
\maketitle

\begin{abstract}
Understanding human visual search behavior is a fundamental problem in vision science and computer vision, with direct implications for modeling how observers allocate attention in location-unknown search tasks. In this study, we investigate the relationship between Gabor-based features and gray-level co-occurrence matrix (GLCM)–based texture features in modeling early-stage visual search behavior. Two feature-combination pipelines are proposed to integrate Gabor and GLCM features for narrowing the region of possible human fixations. The pipelines are evaluated using simulated digital breast tomosynthesis images. Results show qualitative agreement among fixation candidates predicted by the proposed pipelines and a threshold-based model observer. A strong correlation ($r = 0.765$) is observed between GLCM mean and Gabor feature responses, indicating that these features encode related image information despite their different formulations. Eye-tracking data from human observers further suggest consistency between predicted fixation regions and early-stage gaze behavior. These findings highlight the value of combining structural and texture-based features for modeling visual search and support the development of perceptually informed observer models.
\end{abstract}


\section{Introduction}
Human visual search is a fundamental problem in vision science, concerned with how observers allocate attention and extract relevant information from complex visual scenes. In many applied settings, including medical imaging, effective system evaluation and optimization depend on observer performance, since human observers are the ultimate interpreters of image data. Understanding visual search behavior during image interpretation is therefore critical for predicting observer performance and for the development of perceptually informed model observers.

Model observers have been developed as an essential tool for task-based image quality assessment \cite{barrett1993model, Myers:87,rolland1992effect,yao1992predicting,gifford2000channelized,lago2020foveated}. Among various forms of model observers, the Visual Search Model Observer (VSMO) has been introduced to predict observer performance in location-unknown tasks in medical image interpretation by simulating how radiologists search for and detect abnormalities in images using a two-stage search-and-decision process \cite{gifford2013visual,gifford2014efficient,lau2013towards,sen2014assessment,das2011comparison,karbaschi2018assessing,das2015examining,gifford2016visual}. In the first stage of VSMO, the fixation locations are estimated using difference-of-Gaussian (DoG) functions or Gabor functions. These features work as morphological match filters in the spatial domain and as band-pass filters in the frequency domain. Recently, an ideal observer to handle thresholded data in the location-unknown task has been proposed to further narrow down the possible fixation locations\cite{lin2024addition}.

Meanwhile, it has been shown that certain second-order statistical image texture features can predict
signal detection difficulty in tomographic breast images\cite{nisbett2020correlation,kavuri2020relative,kavuri2025examining}. Texture features provide a quantitative description of spatial intensity patterns in images, capturing properties such as granularity, regularity, and local structural variation that are not fully characterized by simple intensity or edge-based measures. Among these, gray-level co-occurrence matrix (GLCM)–based features characterize texture by modeling the joint statistical distribution of gray-level pairs at specified spatial relationships, enabling sensitivity to both local contrast and spatial organization. The GLCM-based features have been shown to enhance the performance of fixation pattern prediction\cite{andrade2024image,andrade2025assessment}.

In this study, two pipelines have been designed to test the effect of combining GLCM–based features and Gabor features to narrow down the region of possible human fixations. The Gaussian mixture model is used in both pipelines to assist in the classification of either GLCM–based features or Gabor features. Simulated digital breast tomosynthesis 2D images are used to test the pipelines. The results are validated by visual exam and correlation calculations.

\section{Materials and Methods}
\label{sec:Materials and Methods}

\subsection{Data Simulation}
\begin{figure}[t]
    \centering
    \begin{subfigure}{0.32\textwidth}
        \centering
        \includegraphics[width=\linewidth]{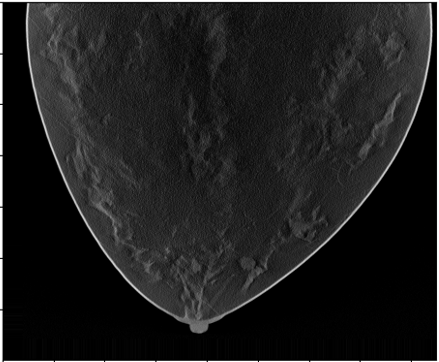}
        \caption{Example of a 2D DBT slice of a breast phantom with density ``mostly fatty''.}
        \label{fig:dbt_fatty}
    \end{subfigure}
    \hfill
    \begin{subfigure}{0.32\textwidth}
        \centering
        \includegraphics[width=\linewidth]{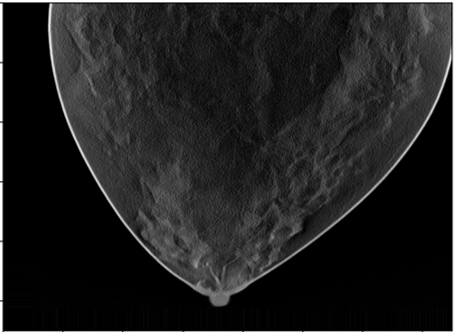}
        \caption{Example of a 2D DBT slice of a breast phantom with density ``scattered areas of dense''.}
        \label{fig:dbt_scattered}
    \end{subfigure}
    \hfill
    \begin{subfigure}{0.32\textwidth}
        \centering
        \includegraphics[width=\linewidth]{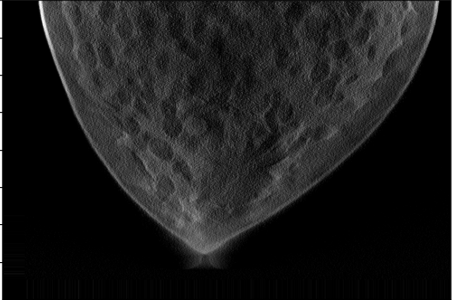}
        \caption{Example of a 2D DBT slice of a breast phantom with density ``heterogeneously dense''.}
        \label{fig:dbt_hetero}
    \end{subfigure}
    \caption{Examples of 2D DBT slices of breast phantoms with different density categories.}
    \label{fig:dbt_density_examples}
\end{figure}

The DBT simulations were generated using the VICTRE pipeline \cite{badano2018evaluation}. We used the VICTRE software tools for phantom generation, X-ray production, and volume reconstruction. The X-ray projection was modeled using the Monte Carlo method based on tissue-radiation interaction physics. Simulated digital breast tomosynthesis (DBT) images were generated using an X-ray source operating at 30 kVp. The acquisition protocol included 25 projection images uniformly distributed over a 50-degree angular range. Image reconstruction was performed using a filtered back projection (FBP) algorithm, producing a voxel size of 0.1 
\( \text{cm}^3 \). We generated breast images for three different breast densities, corresponding to the "mostly fatty", "scattered areas of dense" and "heterogeneously dense" categories described in the BI-RADS report. A spiculated mass (10 mm radius) was created at random locations near terminal duct lobular units.
Examples are shown in Figs.~\ref{fig:dbt_fatty}--\ref{fig:dbt_hetero}

\subsection{Texture Features}

Texture features describe spatial patterns of intensity variation in images and provide information beyond individual pixel values, capturing local structure and contrast that are relevant to visual perception. In medical imaging, such features are commonly used to characterize background complexity and tissue heterogeneity.

In this study, texture information was quantified using features derived from the gray-level co-occurrence matrix (GLCM). For a given image region, the GLCM is defined as the normalized joint probability of observing two pixels with gray levels $i$ and $j$ separated by a specified spatial offset $\boldsymbol{\delta}$. Let $N(i,j \mid \boldsymbol{\delta})$ denote the number of pixel pairs with gray levels $i$ and $j$ at offset $\boldsymbol{\delta}$, where $i,j \in \{0,\ldots,G-1\}$. The normalized GLCM is given by
\begin{equation}
P(i,j \mid \boldsymbol{\delta}) = \frac{N(i,j \mid \boldsymbol{\delta})}{\sum_{i=0}^{G-1}\sum_{j=0}^{G-1} N(i,j \mid \boldsymbol{\delta})}.
\tag{1}
\end{equation}

The first texture feature used was the GLCM mean, defined as the expected gray level of the reference pixel,
\begin{equation}
\mathrm{GLCM}_{\text{mean}} = \sum_{i=0}^{G-1}\sum_{j=0}^{G-1} i \, P(i,j \mid \boldsymbol{\delta}).
\tag{2}
\end{equation}

The second texture feature was the GLCM contrast, which measures the degree of local intensity variation between neighboring pixels,
\begin{equation}
\mathrm{GLCM}_{\text{contrast}} = \sum_{i=0}^{G-1}\sum_{j=0}^{G-1} (i - j)^2 \, P(i,j \mid \boldsymbol{\delta}).
\tag{3}
\end{equation}

These features provide complementary descriptions of texture, with the GLCM mean capturing average gray-level structure and the GLCM contrast capturing local intensity differences that are relevant to visual search behavior.

\subsection{Gabor Features}
The 2D Gabor function is an essential tool for localized frequency and orientation analysis in image processing. It is defined as follows:

\begin{equation}
\begin{split}
G(x, y) = \exp\left[-4(\ln 2)\frac{(x - x_0)^2 + (y - y_0)^2}{W_s^2}\right] \\
\cdot \cos\left[2\pi f_c \left((x - x_0)\cos(\theta) + (y - y_0)\sin(\theta)\right) + \phi\right]
\end{split},\tag{4}
\end{equation}
where \( (x, y) \) are the spatial coordinates of the image, and \( (x_0, y_0) \) represent the center of the Gabor filter in the spatial domain. The parameter \( W_s \) denotes the width of the Gaussian envelope, which controls the spatial spread of the filter. The term \( f_c \) specifies the central frequency of the sinusoidal component, determining the filter’s frequency sensitivity. The orientation of the sinusoidal wave is given by \( \theta \), allowing the filter to be tuned to specific edge directions, while \( \phi \) is the phase offset, shifting the cosine wave to match particular features in the image. These parameters were set to fit the size of images and targets.
\subsection{Gaussian Mixture Model}
A Gaussian mixture model (GMM) is an unsupervised clustering method that models data as a combination of multiple Gaussian components. It assigns data points to multiple clusters to identify underlying group structure in feature spaces without requiring labeled data.
\subsection{The GLCM-Gabor Pipeline A}
Two pipelines to combine GLCM and Gabor features are explored, the first one will be referred as GLCM-Gabor pipeline, or pipeline A.
Pipeline A starts by calculating GLCM mean and GLCM contrast over the entire image. The quantization and window size parameter for GLCM calculation is set to 128 and 100. Based on these two features, the pixels are clustered into 5 groups by GMM, including one group representing the background. After clustering, a mask is generated using the cluster that contains the region of the lesion.
Four Gabor features are also calculated over the entire image using cross correlation;  this is the first stage of the visual search model observer\cite{gifford2013visual,gifford2014efficient,lau2013towards,sen2014assessment,das2011comparison,karbaschi2018assessing,das2015examining}. A reversed-watershed algorithm is used to select local maxima in the four feature maps. These local maxima are considered initial fixation candidates. These candidates are then further screened by applying the mask from previous texture feature cluster results. The remaining candidates are considered the final estimation of fixation locations.
\subsection{The Gabor-GLCM Pipeline B}
In pipeline B, the first step is to calculate the feature map with the four Gabor features using cross-correlation. Initial candidates are selected by finding the regional maxima. Then calculate the GLCM mean and GLCM contrast only on these candidates locations. After that, the candidates are clustered by GMM using six total features (four Gabor features and two GLCM-based features). The resulting candidates in the group with the lesion are considered the final estimation of fixation locations. 
\subsection{Candidates Selection Based on Thresholded Data}
As a comparison, the candidates selected based on thresholded data are also calculated. This is part of the progress of a model observer designed for thresholded data. This model is shown to predict human performance well \cite{lin2024addition}. The same four Gabor features are used in this method. The thresholded data is acquired by setting a lower threshold to keep only the candidates with larger feature values.
\section{Results}
Results of applying both pipelines on a 2D DBT slice from a fatty density phantom are shown here, more results could be found in section ~\ref{sec:supp}.  Fig.~\ref{fig:fatty1_pipeline1_segs} shown the clustering result using GLCM mean and GLCM contrast from pipeline A. A mask is produced from the group that contained the lesion as  Fig.~\ref{fig:fatty1_pipeline1_mask}. The final predicted candidates are shown in  Fig.~\ref{fig:fatty1_pipeline1_result}.

The initial candidates resulting from pipeline B is shown in Fig.~\ref{fig:fatty1_pipeline2_init}. Fig. ~\ref{fig:fatty1_pipeline2_result} shown the cluster based on all features calculated only on the candidates locations. Fig. ~\ref{fig:fatty1_pipeline2_result_2} shown the final estimation of pipeline B.

Visual inspection indicates qualitative agreement among both candidates location results from pipeline A, B, and the thresholded model, as shown in Fig. ~\ref{fig:thr_result}. 

\begin{center}
\begin{minipage}{0.48\textwidth}
    \centering
    \includegraphics[height=5.5cm]{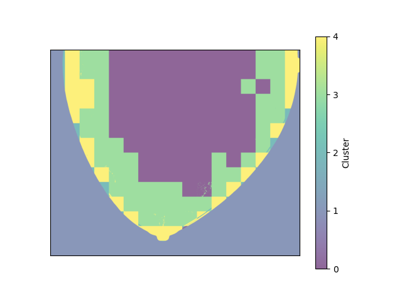}
    \captionof{figure}{Segmentation based on texture features.}
    \label{fig:fatty1_pipeline1_segs}
\end{minipage}
\hfill
\begin{minipage}{0.48\textwidth}
    \centering
    \includegraphics[height=5.5cm]{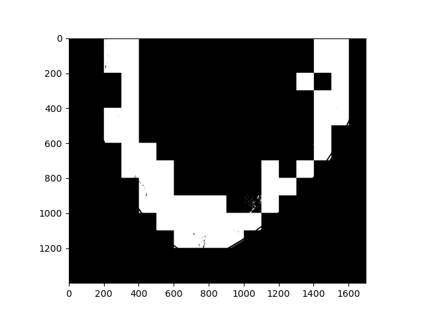}
    \captionof{figure}{Mask based on texture features.}
    \label{fig:fatty1_pipeline1_mask}
\end{minipage}
\end{center}

\begin{center}
\begin{minipage}{0.48\textwidth}
    \centering
    \includegraphics[height=5.5cm]{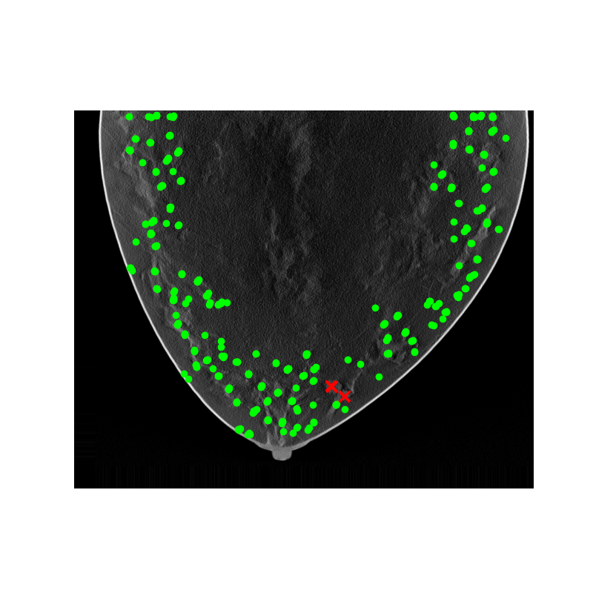}
    \captionof{figure}{Clustered candidates based on texture features.}
    \label{fig:fatty1_pipeline1_result}
\end{minipage}
\hfill
\begin{minipage}{0.48\textwidth}
    \centering
    \includegraphics[height=5.5cm]{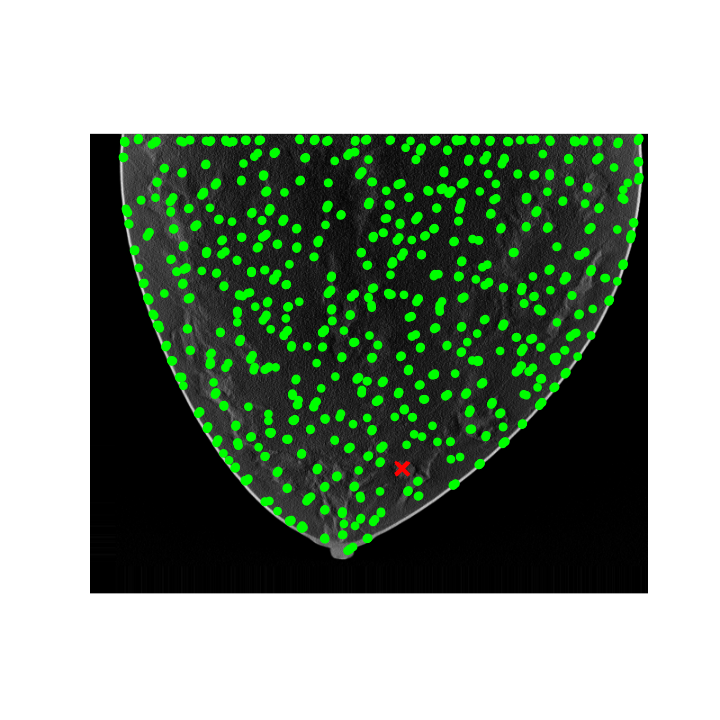}
    \captionof{figure}{Initial candidates from pipeline B.}
    \label{fig:fatty1_pipeline2_init}
\end{minipage}
\end{center}

\begin{center}
\begin{minipage}{0.48\textwidth}
    \centering
    \includegraphics[height=5.5cm]{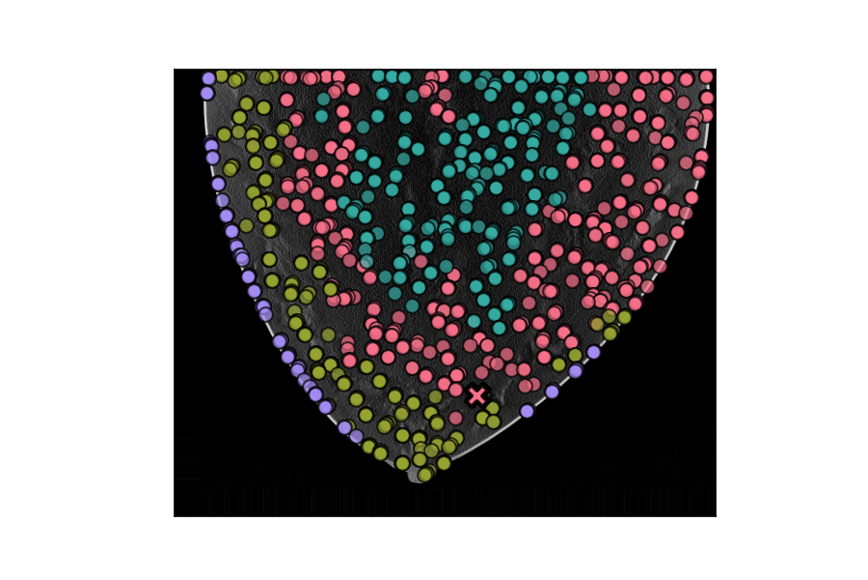}
    \captionof{figure}{Initial candidates clustered by all features.}
    \label{fig:fatty1_pipeline2_result}
\end{minipage}
\hfill
\begin{minipage}{0.48\textwidth}
    \centering
    \includegraphics[height=5.5cm]{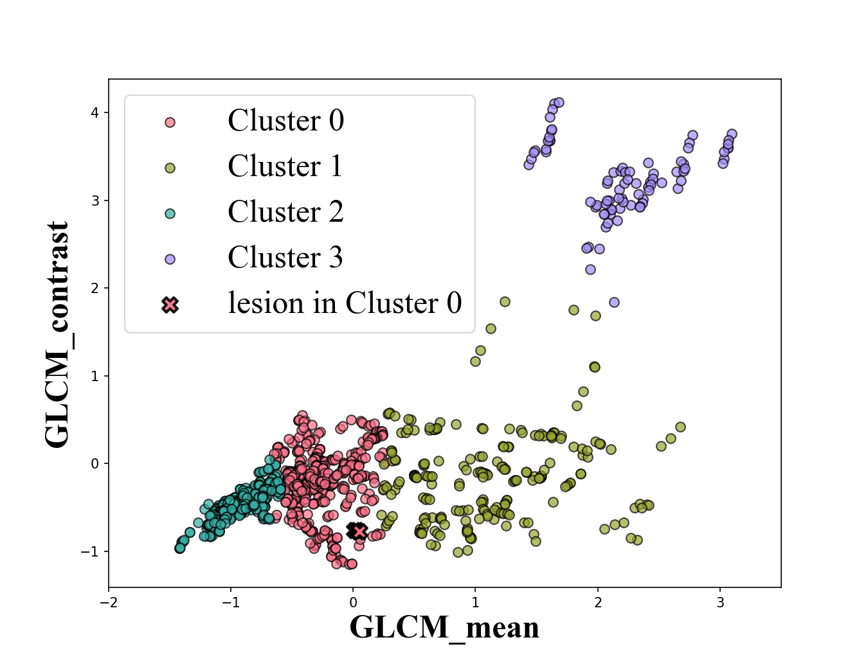}
    \captionof{figure}{Scattered plots of texture feature calculated on the initial candidates.}
    \label{fig:fatty1_pipeline2_plot}
\end{minipage}
\end{center}

\begin{center}
\begin{minipage}{0.48\textwidth}
    \centering
    \includegraphics[height=5.5cm]{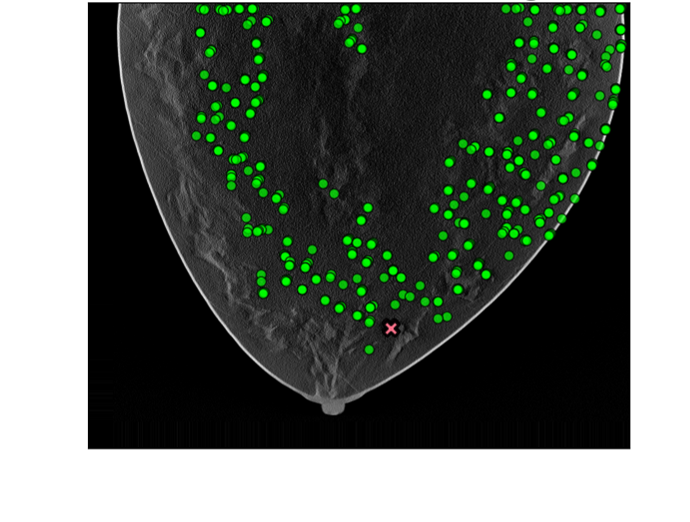}
    \captionof{figure}{Clustered candidates from Gabor features.}
    \label{fig:fatty1_pipeline2_result_2}
\end{minipage}
\hfill
\begin{minipage}{0.48\textwidth}
    \centering
    \includegraphics[height=5.5cm]{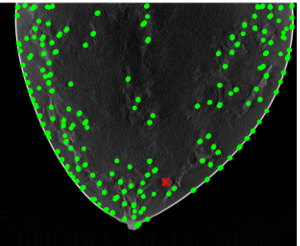}
    \captionof{figure}{Candidates estimated from model observer on thresholded data.}
    \label{fig:thr_result}
\end{minipage}
\end{center}

\begin{figure}[H]
\begin{center}
\begin{tabular}{c}
\includegraphics[height=5.5cm]{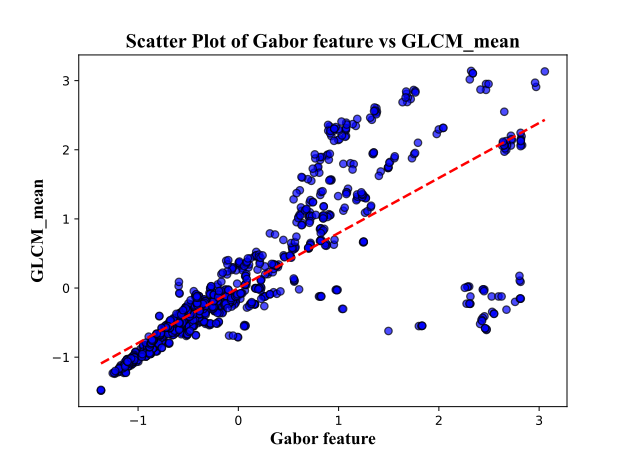}
\end{tabular}
\end{center}
\caption 
{ \label{fig:corr}
Correlation between Gabor and texture features. } 
\end{figure} 
\section{Discussion}
\begin{figure}[t]
    \begin{subfigure}{0.5\textwidth}
        \centering
        \includegraphics[width=\linewidth]{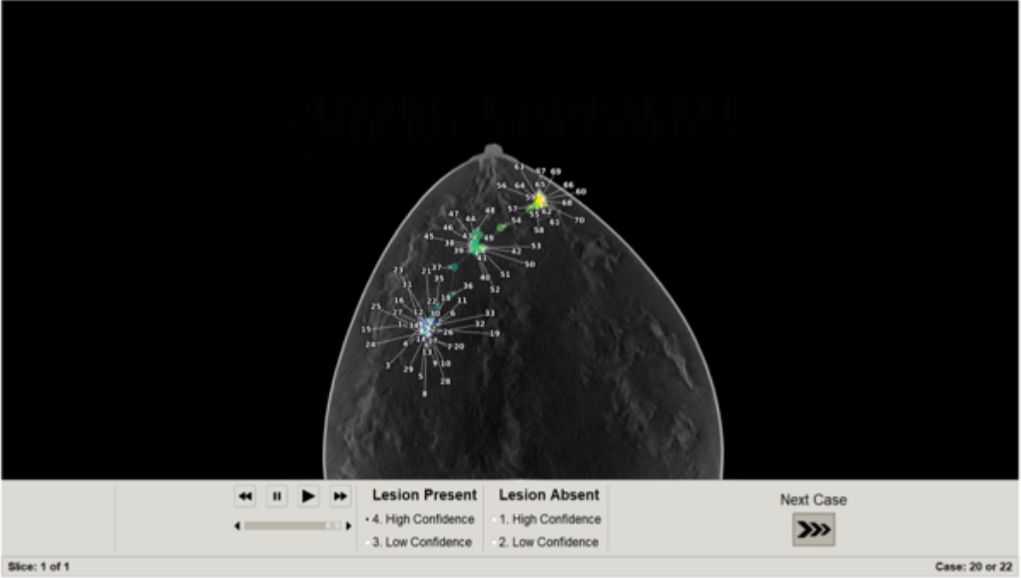}
        
        \label{fig:ob1}
    \end{subfigure}
    \hfill
    \begin{subfigure}{0.5\textwidth}
        \centering
        \includegraphics[width=\linewidth]{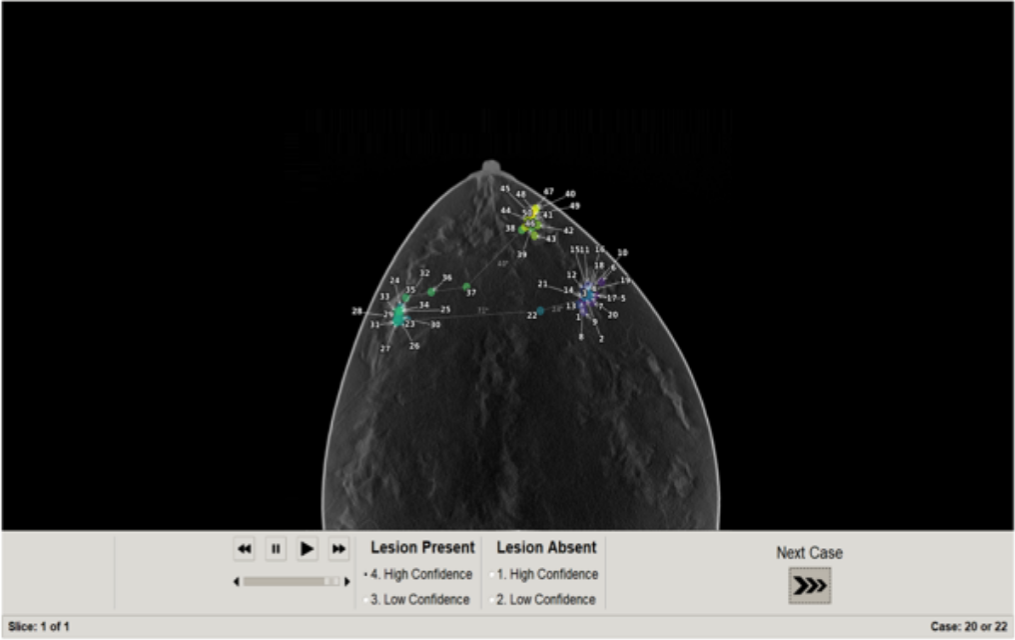}
        
        \label{fig:ob2}
    \end{subfigure}
    \caption{Early stage search gaze points of observer 1 and observer 2.}
    \label{fig:ob12}
\end{figure}
This study investigates the relationship between Gabor-based features and GLCM-based texture features in modeling early-stage visual search behavior. These feature classes provide complementary descriptions of image content: Gabor features capture localized orientation- and frequency-selective structure, while GLCM features characterize statistical properties of local intensity patterns that describe image texture. The computation of Gabor features generate a same size feature map as the image, so it provide pixel level result, while the result of texture features are patches that defined by the window size parameter.  Examining their relationship offers insight into how different image representations may jointly influence fixation behavior during visual search.

The findings can be interpreted in the context of the global ``gist'' theory of medical image perception\cite{evans2013gist}. Gist theory proposes that observers rapidly extract a global impression of an image, reflecting overall texture and structural organization, prior to focused inspection of local regions. Within this framework, GLCM features may be viewed as quantitative descriptors of coarse texture statistics that contribute to gist formation, whereas Gabor features capture localized structural cues that may guide subsequent attentional deployment.

A key result of this study is the strong correlation observed between the GLCM mean feature and the Gabor-based feature responses, with a correlation coefficient of 0.765, see Fig.~\ref{fig:corr}. This level of correlation indicates that, despite their different formulations, these features encode related information about image structure. Importantly, this relationship suggests that local orientation-sensitive responses captured by Gabor filters are systematically influenced by underlying texture statistics represented by the GLCM mean. We also examined our result with human eye-tracking (Tobii pro, 120hz) result, see Fig.~\ref{fig:ob12}. The average recorded gaze point locations lie within the region of results from pipelines A and B.

Overall, the results indicate that combining Gabor and GLCM features provides a more complete characterization of image properties relevant to early fixation selection in visual search tasks. This integrated representation has broader implications for understanding and modeling attention allocation in complex visual scenes, and it also informs the development of perceptually grounded observer models for applications involving location uncertainty, including but not limited to imaging system evaluation.

In the future, more human eye-tracking data will be used to compare with the models. We are also planning to examine a much larger scale of features to answer the question of which features guide visual attention.

\section{Conclusion}

This study examined the relationship between Gabor-based features and GLCM-based texture features for modeling early-stage visual search behavior in medical images. Two pipelines were developed to integrate these feature classes within a VSMO framework and evaluated using simulated DBT images. Both pipelines produced fixation candidate estimates that were qualitatively consistent with each other and with a threshold-based model observer.

A key finding was the strong correlation observed between the GLCM mean feature and Gabor feature responses, indicating that local orientation-sensitive structure and underlying texture statistics are closely related in the images studied. This result suggests that texture information captured by second-order statistics may influence or constrain local feature responses that guide early fixation selection. The observed agreement between predicted fixation regions and preliminary human eye-tracking data further supports the perceptual relevance of the proposed feature combinations.

Overall, the results demonstrate that integrating Gabor and GLCM features provides a more comprehensive representation of image properties relevant to early visual search. This integrated approach has broader implications for modeling attention allocation and search behavior in complex visual environments, and it also informs the development of perceptually grounded observer models for applications involving location uncertainty. Future work will include validation with larger eye-tracking datasets and the exploration of additional feature families to further clarify the mechanisms guiding visual attention across visual search tasks.

\section*{Acknowledgments}
This work was partially supported by funding from the (NIBIB) grant NIBIB R01 EB03246 \& (NIBIB) R01 EB029761,
NSF 1652892 and the US Department of Defense (DOD) Congressionally Directed Medical Research Program (CDMRP)
Breakthrough Award BC151607

\bibliographystyle{unsrt}  
\bibliography{references}  
\clearpage
\section{Supplements Material}
\label{sec:supp}
\setcounter{figure}{0}
\begin{center}
\begin{minipage}{0.48\textwidth}
    \centering
    \includegraphics[height=5.5cm]{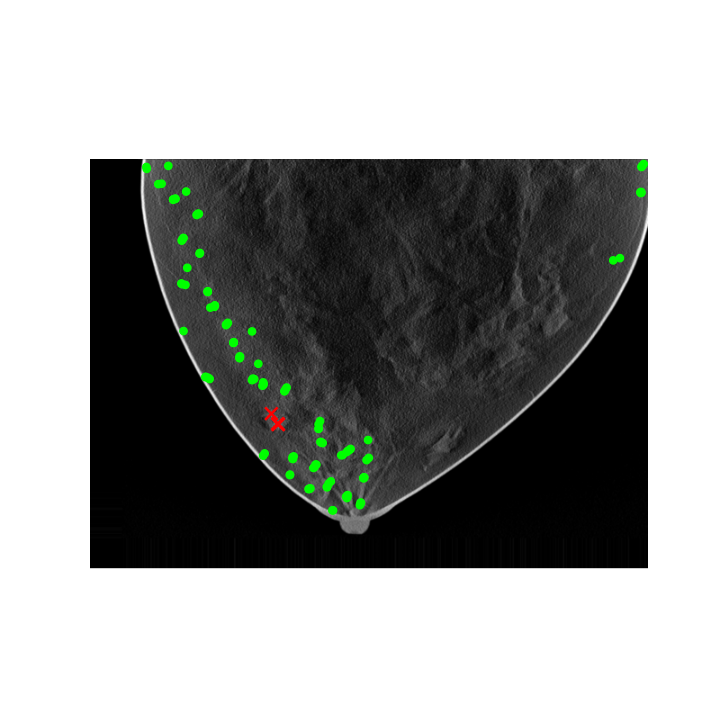}
    \captionof{figure}{"Scattered" DBT pipeline A result}
    \label{fig:Scattered_1}
\end{minipage}
\hfill
\begin{minipage}{0.48\textwidth}
    \centering
    \includegraphics[height=5.5cm]{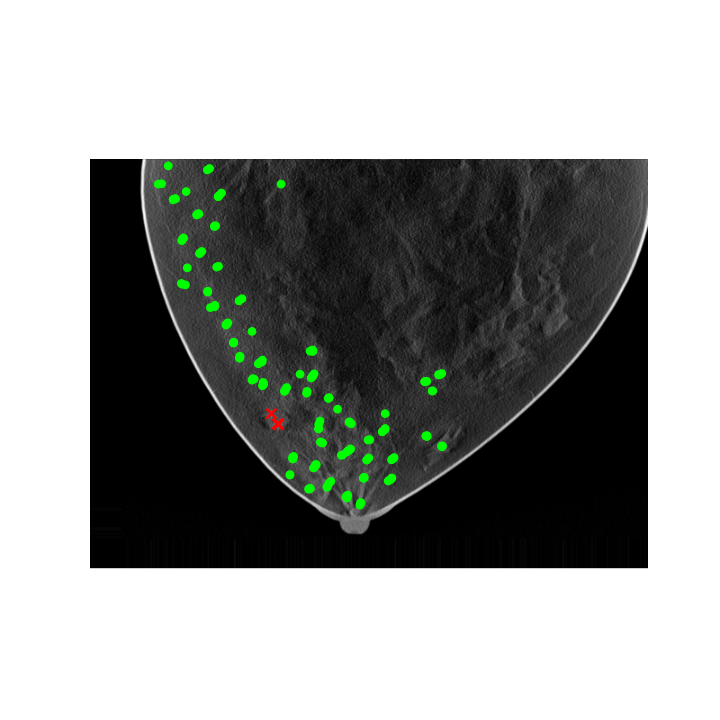}
    \captionof{figure}{"Scattered" DBT pipeline B result}
    \label{fig:Scattered_2}
\end{minipage}
\end{center}
\begin{center}
\begin{minipage}{0.48\textwidth}
    \centering
    \includegraphics[height=5.5cm]{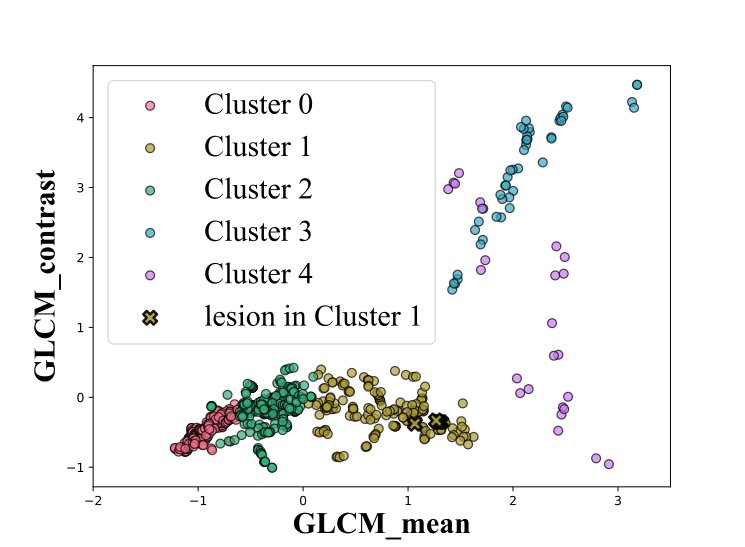}
    \captionof{figure}{"Scattered" density DBT scattered plot}
    \label{fig:Scattered_3}
\end{minipage}
\end{center}

\begin{center}
\begin{minipage}{0.48\textwidth}
    \centering
    \includegraphics[height=5.5cm]{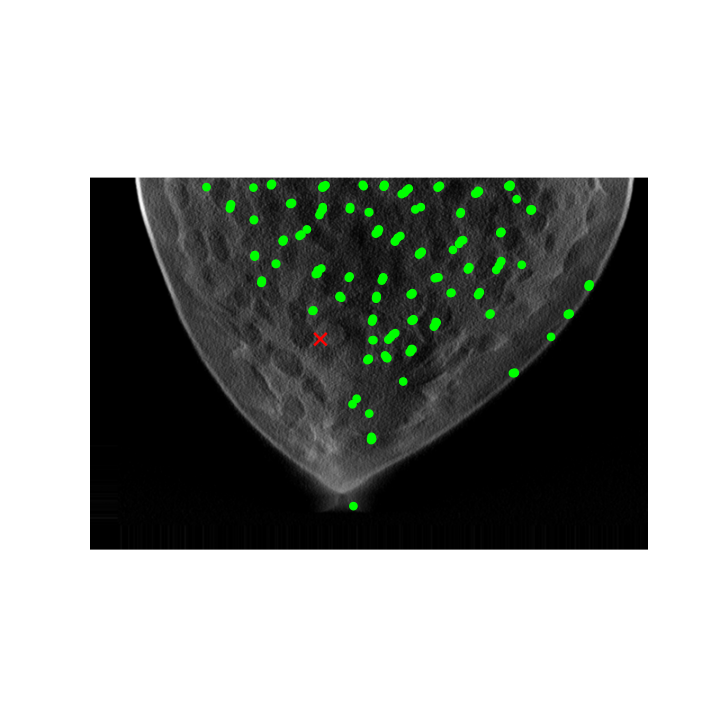}
    \captionof{figure}{"Heterogeneously dense" DBT pipeline A result}
    \label{fig:het_1}
\end{minipage}
\hfill
\begin{minipage}{0.48\textwidth}
    \centering
    \includegraphics[height=5.5cm]{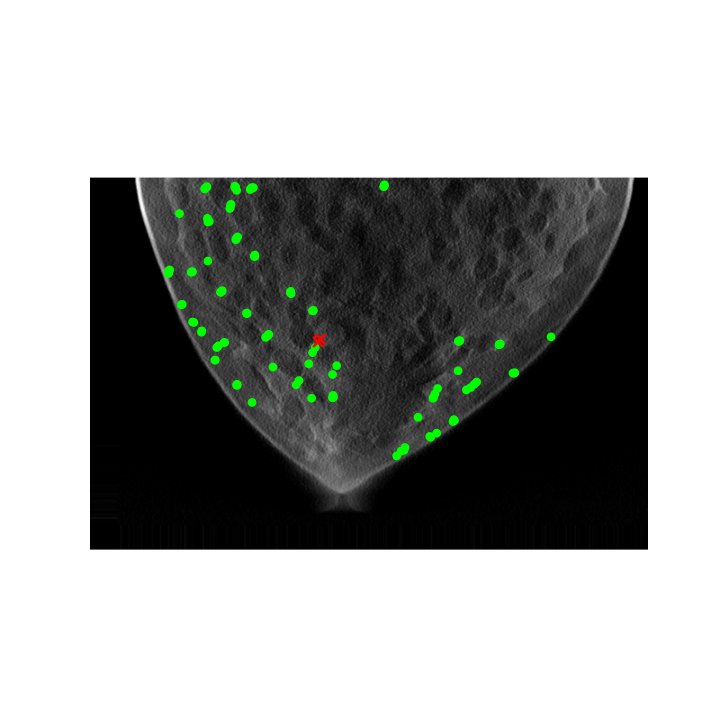}
    \captionof{figure}{"Heterogeneously dense" DBT pipeline B result}
    \label{fig:het_2}
\end{minipage}
\end{center}

\begin{center}
\begin{minipage}{0.48\textwidth}
    \centering
    \includegraphics[height=5.5cm]{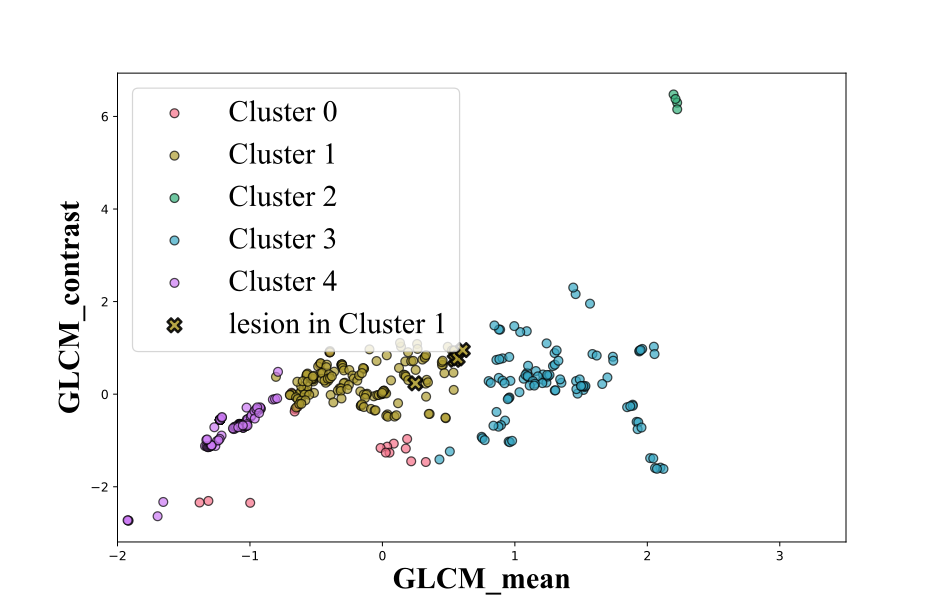}
    \captionof{figure}{"Heterogeneously dense" density DBT scattered plot}
    \label{fig:het_3}
\end{minipage}
\end{center}

\end{document}